%% file: TiSSiLe.tex
\documentclass{article}

\usepackage{times}
\usepackage{graphicx} 
\usepackage{subfigure} 
\usepackage[table]{xcolor}

\usepackage{natbib}

\usepackage{algorithm}
\usepackage{algorithmic}

\usepackage{amsmath}

\usepackage{tikz}
\usetikzlibrary{arrows,automata}
\usetikzlibrary{positioning}
\usepackage{xspace}

\usepackage{multirow}

\usepackage[draft]{hyperref}

\definecolor{gray}{gray}{0.7}

\usepackage[accepted]{icml2016}

\icmltitlerunning{Modeling Time Series Similarity with Siamese Recurrent Networks}

\begin{document} 

\twocolumn[
\icmltitle{Modeling Time Series Similarity with Siamese Recurrent Networks}

\icmlauthor{Wenjie Pei}{w.pei-1@tudelft.nl}
\icmlauthor{David M.J. Tax}{d.m.j.tax@tudelft.nl}
\icmladdress{Delft University of Technology, Mekelweg 4, 2628 CD Delft, THE NETHERLANDS}
\icmlauthor{Laurens van der Maaten}{lvdmaaten@fb.com}
\icmladdress{Facebook AI Research, 770 Broadway 8th Floor, New York NY 10003, USA}



\icmlkeywords{time series, representation learning, recurrent neural networks}

\vskip 0.3in
]

\newcommand {\commenting}[1]{\textcolor[rgb]{1,0,0} {#1}}
\newcommand {\emphing}[1]{\textcolor[rgb]{0,0,1} {#1}}
\newcommand {\SRN}{SRN\xspace}
\newcommand {\simil}{s}

\begin{abstract} 
\input{abstract}
\end{abstract}

\section{Introduction}
\input{introduction}

\section{Related Work}
\input{related_work}

\section{Siamese Recurrent Networks}
\input{models}

\section{Experiments}
\input{experiments}

\section{Conclusions}
\input{conclusions}

\section*{Acknowledgments}
This work was supported by AAL SALIG++.

\nocite{langley00}

\bibliography{TiSSiLe}
\bibliographystyle{icml2016}

\end{document}

%% file: abstract.tex

Traditional techniques for measuring similarities between time series are based on handcrafted similarity measures, whereas more recent learning-based approaches cannot exploit external supervision. We combine ideas from time-series modeling and metric learning, and study \emph{siamese recurrent networks} (SRNs) that minimize a classification loss to learn a good similarity measure between time series. Specifically, our approach learns a vectorial representation for each time series in such a way that similar time series are modeled by similar representations, and dissimilar time series by dissimilar representations. Because it is a similarity prediction models, SRNs are particularly well-suited to challenging scenarios such as signature recognition, in which each person is a separate class and very few examples per class are available. We demonstrate the potential merits of SRNs in within-domain and out-of-domain classification experiments and in one-shot learning experiments on tasks such as signature, voice, and sign language recognition.

%% file: introduction.tex
Successful classification, verification, or retrieval of time series requires the definition of a good similarity measure between time series. Classical approaches to time-series analysis handcraft such similarity measures \citep{vintsyuk68,sakoe_DTW}, which limits their ability to incorporate information on the relative scale of features in the similarity measure. Other approaches use unsupervised learning in order to define the similarity measure~\citep{rabiner89,FisherKernel}, which has the disadvantage that it cannot exploit class label information in determining which features are most relevant for the underlying similarity structure. 

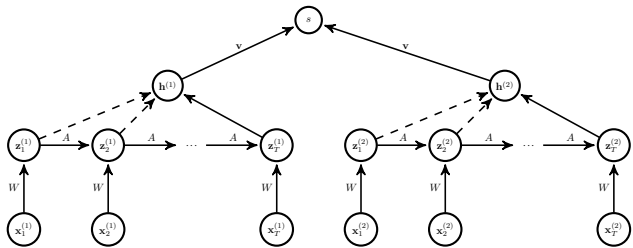
\begin{figure}[t]
\centering
\begin{tikzpicture}[scale=0.35, every node/.style={scale=0.40},->,>=stealth',shorten >=1pt,auto,node distance=2.8cm,semithick]
\tikzstyle{every state}=[fill=white,draw=black,thick]
\node[state] (z1) {$\mathbf{z}_1^{(1)}$};
\node[state] (z2) [right of=z1] {$\mathbf{z}_2^{(1)}$};
\node[state,draw=none] (z3) [right of=z2] {\dots};
\node[state] (zT) [right of=z3] {$\mathbf{z}_T^{(1)}$};
\path (z1) edge node[above] {$A$} (z2);
\path (z2) edge node[above] {$A$} (z3);
\path (z3) edge node[above] {$A$} (zT);
\node[state] (H1) [above right of=z2] {$\mathbf{h}^{(1)}$};
\path[dashed] (z1) edge node[left] {} (H1);
\path[dashed] (z2) edge node[left] {} (H1);
\path (zT) edge node[left] {} (H1);
\node[state] (x1) [below of=z1] {$\mathbf{x}_1^{(1)}$};
\node[state] (x2) [below of=z2] {$\mathbf{x}_2^{(1)}$};
\node[state] (xT) [below of=zT] {$\mathbf{x}_T^{(1)}$};
\path (x1) edge node[left] {$W$} (z1);
\path (x2) edge node[left] {$W$} (z2);
\path (xT) edge node[left] {$W$} (zT);

\node[state] (z4) [right of=zT] {$\mathbf{z}_1^{(2)}$};
\node[state] (z5) [right of=z4] {$\mathbf{z}_2^{(2)}$};
\node[state,draw=none] (z6) [right of=z5] {\dots};
\node[state] (zT) [right of=z6] {$\mathbf{z}_T^{(2)}$};
\path (z4) edge node[above] {$A$} (z5);
\path (z5) edge node[above] {$A$} (z6);
\path (z6) edge node[above] {$A$} (zT);
\node[state] (H2) [above right of=z5] {$\mathbf{h}^{(2)}$};
\path[dashed] (z4) edge node[left] {} (H2);
\path[dashed] (z5) edge node[left] {} (H2);
\path (zT) edge node[left] {} (H2);
\node[state] (x4) [below of=z4] {$\mathbf{x}_1^{(2)}$};
\node[state] (x5) [below of=z5] {$\mathbf{x}_2^{(2)}$};
\node[state] (xT) [below of=zT] {$\mathbf{x}_T^{(2)}$};
\path (x4) edge node[left] {$W$} (z4);
\path (x5) edge node[left] {$W$} (z5);
\path (xT) edge node[left] {$W$} (zT);

\node[state] (O) [above right = 0.6cm and 1.6cm of H1] {$\simil$};
\path (H1) edge node[above] {$\mathbf{v}$} (O);
\path (H2) edge node[above] {$\mathbf{v}$} (O);
\end{tikzpicture}

\caption{Graphical representation of the Siamese Recurrent Network
(\SRN). For the SRN-L model, the feature representations $\mathbf{h}$ are obtained by taking the hidden unit activations at the last timestep, $\mathbf{z}_T$ (solid line). For the SRN-A model, the feature representations $\mathbf{h}$ are obtained by averaging the hidden unit activations $\mathbf{z}$ over all timesteps (solid and dashed lines). The SRN outputs a scalar similarity measure $\simil$.}
\label{fig:SRNgraph}
\end{figure}

In this paper, we study a novel model for time-series analysis that \emph{learns} a similarity measure over pairs of time series in a \emph{supervised} manner. The proposed model combines ideas from metric learning with that of learning embeddings for time series using recurrent networks. The model takes as input two time series, which are both processed by the same recurrent network to produce a representation for each of time series. The similarity between the time series is defined as a weighted inner product between the resulting representations. All parameters of the model are learned jointly by minimizing a classification loss on pairs of similar and dissimilar time series. We refer to the resulting model as \emph{siamese recurrent network} (SRN). The structure of the SRN is illustrated in Figure~\ref{fig:SRNgraph}. We evaluate the performance of two variants of the SRN in within-domain classification and out-of-domain classification experiments representing a range of different machine-learning tasks.

The model we study in this paper is of particular interest in challenging learning settings in which the number of classes is large and the number of training examples per class is limited. An example of such a setting is an online signature verification task. Here each person who provided one or more signatures is considered to be a separate class, and the number of training examples per person is extremely limited. Such a task may benefit from sharing parameters between classes by learning a global similarity measure over the set of all pairs of time series, which is what the SRN does. We perform one-shot learning experiments to illustrate the potential merits of the global similarity measure over time series learned by our models. 


%% file: related_work.tex
Traditional approaches to measuring time-series similarity such as dynamic time warping (DTW; \citet{vintsyuk68,sakoe_DTW}) use handcrafted similarity measures that are not adapted to the observed data distribution. This shortcoming  was addressed by the introduction of similarity measures that first fit a generative model to the data, such as Fisher, TOP, marginalized, and product-probability kernels~\citep{FisherKernel,top_kernel,product_kernel,marginalized_kernels}. In particular, Fisher kernels have seen widespread adoption in computer vision \citep{perronnin2010}. While these methods benefit from modeling the data distribution before the computation of pairwise similarities, they are limited in that they cannot exploit available supervised class or similarity information, which may hamper their performance in classification problems. By contrast, the time-series similarity approach we study in this work is based on \emph{supervised learning}. It combines ideas from modeling time series using recurrent networks with those from metric learning. We discuss related work on both topics separately below.

\textbf{Recurrent networks} learn a representation for each timestep that is influenced by both the observation at that time step and by the representation in the previous timestep \citep{werbos88,schmidhuber89}. The recurrent nature of the models equips them with a memory that is capable of preserving information over time. This has made them popular for tasks such as language~\citep{rnn_language, Vinyals2015}, handwriting \citep{graves13}, and image generation \citep{Theis2015c}, and music prediction \citep{bengio13}. SRNs employ a pair of standard recurrent networks, the parameters of which are shared between the two networks. It differs from prior work in the loss that it minimizes: instead of minimizing a ``generative'' loss such as negative log-likelihood, it minimizes a loss that encourages representations to be close together for similar time series and far apart for dissimilar time series.

\textbf{Metric learning} techniques learn a similarity measure on data that lives in a vectorial space. While several studies have explored learning non-linear ``metrics'' by backpropagating pairwise losses through feedforward networks \citep{bromley93,chopra05,SalHinton07,koch2015,min10,hadsell06,hu14}, most prior work on metric learning focuses on learning Mahalanobis metrics; prominent examples of such studies include \citet{goldberger04,weinberger09distance,davis07}; and \citet{xing02}. Our work is most similar to latent coincidence analysis (LCA; \citet{der12}) in terms of the loss it is minimizing, but it differs substantially from LCA in that it backpropagates the loss through the recurrent network that is modeling the time series.

%% file: models.tex
\label{models_section}

A time-series similarity model produces a single similarity value for each input pair of time series (with potentially different lengths).  Similarly to a siamese network, our time-series similarity model employs two neural networks that share their network parameters in order to extract comparable hidden-unit representations from the inputs. The resulting hidden-unit representations are compared to compute the similarity between the two time series. The parameters of the neural networks and the comparison function are learned jointly in a supervised manner to predict whether two time series are similar or not.  We use recurrent networks as the basis for our siamese architecture, leading to the siamese recurrent network (SRN) depicted in Figure~\ref{fig:SRNgraph}.  The advantage of using recurrent networks is that they allow our model (1) to extract relevant features for the similarity computation and (2) to remember these relevant features over time when needed. The resulting features have the same size irrespective of the time series length.

Suppose we are given two time series $X^{(1)}\!=\!\left\{\mathbf{x}^{(1)}_1, \dots,
\mathbf{x}^{(1)}_{T_1}\right\}$ and $X^{(2)}\!=\!\left\{\mathbf{x}^{(2)}_1, \dots,
\mathbf{x}^{(2)}_{T_2}\right\}$ whose lengths are respectively $T_1$ and
$T_2$.
The hidden-unit representations
$\mathbf{z}_t^{(1)}$ and $\mathbf{z}_t^{(2)}$ in the SRN model are defined as:
\begin{align}
   \mathbf{z}_{t}^{(i)} = \mathrm{g}\left(\mathbf{W} \mathbf{x}_{t}^{(i)} + \mathbf{A} \mathbf{z}_{t-1}^{(i)} + \mathbf{b}\right).
\label{eqn:hidden_unit}
\end{align} 
We use a rectified linear unit (ReLU) function $\mathrm{g} (x) = \max(0, x)$ as this
activation function eliminates potential vanishing-gradient problems.

The hidden-unit representations obtained from the two sub-networks for
the corresponding input time series, $\mathbf{h}^{(1)}$ and
$\mathbf{h}^{(2)}$, are combined to compute the SRN's prediction for the
similarity of the two time series. We consider two approaches for
comparing hidden-unit representations.

In the first approach, the
element-wise product between the hidden representations on the last time
steps $T_1$ and $T_2$ is computed and the output is a weighted sum of the resulting
products. This approach encourages the recurrent networks to
remember relevant features over time, thereby making these features
available for the final similarity computation.

In the second approach,
all the hidden-unit representations for each of the two time series are
averaged over time to construct a single feature representation for both
time series, and the resulting feature representations are combined in
the same way as before to compute the time-series similarity. This
approach removes the burden on the recurrent networks to memorize
all important features over time, but may potentially pollute the
time-series features by averaging over time. 

Mathematically, the two
approaches compute the following latent representations
$\mathbf{h}$ for each time series:
\begin{itemize}
   \item The \textbf{\SRN -L} (last timestep) model:
\begin{align}
   \mathbf{h}^{(i)}= \mathbf{h}\left(X^{(i)}\right) = \mathbf{z}^{(i)}_T.
\label{eqn:SRNs_L}
\end{align}     
The recurrent connections in recurrent networks allow it to
memorize the previous inputs in the hidden states in a recursive way.
Consequently, the hidden units in the last time step should be able to store the information accumulated in the time domain for the whole time series. Therefore, we conjecture it is capable of modeling the entire time series.   
\item The \textbf{\SRN -A} (average) model:
\begin{align}
   \mathbf{h}^{(i)} = \mathbf{h}\left(X^{(i)}\right) = \frac{1}{T} \sum_{t=1}^{T} \mathbf{z}^{(i)}_t.
\label{eqn:SRNs_A}
\end{align}
By averaging the hidden units $\mathbf{z}$ over time, this model treats the information of each time step equally and avoids the potential memory-vanishing problem whilst still considering the temporal information in the previous time steps when computing hidden-unit representations.
\end{itemize}

Denoting the latent representations obtained from the two recurrent networks as
$\mathbf{h}^{(1)}$ and $\mathbf{h}^{(2)}$, the SRN model defines the similarity of
the two time series as:
\begin{align}
\simil\left(X^{(1)}, X^{(2)}\right) =
\frac{1}{1 + e^{- \mathbf{v}^\top \left[\mathrm{diag} \left(
\mathbf{h}^{(1)} {\mathbf{h}^{(2)}}^\top   \right) \right] + c }}.
\end{align}

Herein, the similarity between two time series is defined as a weighted inner
product between the latent representations $\mathbf{h}^{(1)}$ and
$\mathbf{h}^{(2)}$. Such similarity measures between hidden-units activations have previously been used as part of attention mechanisms in speech recognition \citep{chorowski14}, machine translation \citep{bahdanau14}, and handwriting generation \citep{graves13}.

\subsection{Parameter Learning}

Suppose we are given a training set $\mathcal{T}$ containing two sets of in total $N$
pairs of time series, a set with pairs of similar time series
$\mathcal{S}$ and a set with pairs of dissimilar time series $\mathcal{D}$.
We learn all parameters $\Theta = \{\mathbf{A}, \mathbf{W}, \mathbf{v}, c, \mathbf{b}\}$ of the SRN jointly by minimizing the binary cross-entropy of predicting to which set each pair of time series belongs with respect to the parameters. This is equivalent to maximizing the conditional log-likelihood of the training data:

\begin{equation}
\begin{split}
\mathcal{L}(\Theta; \mathcal{T}) =  -\bigg[ &\sum_{(n_1, n_2) \in \mathcal{S}} \log \simil\left(X^{(n_1)}, X^{(n_2)}\right) \\
+  &\sum_{(n_1, n_2) \in \mathcal{D}} \log \left(1- \simil\left(X^{(n_1)}, X^{(n_2)}\right) \right) \bigg],
\label{eqn:Loss}\nonumber
\end{split}
\end{equation}  

where $n_1$ and $n_2$ indicate the indices of the first and second time series in a training pair.
The loss function is backpropagated through both recurrent networks (the weights of which are shared) using a variant of the backpropagation through time algorithm \cite{werbos88} with gradient clipping between $-5$ and $5$ \citep{bengio13}.

The sets $\mathcal{S}$ and $\mathcal{D}$ of similar and
dissimilar time series can be constructed in various ways, for instance,
by asking human annotators for similarity judgements. When class labels
$y_n$ are available for time series $X^{(n)}$, the sets can be defined as $\mathcal{S}\!=\!\{(n_1, n_2)\!: y_{n_1}\!=\!y_{n_2} \}$ and $\mathcal{D}\!=\!\{(n_1, n_2)\!: y_{n_1}\!\neq\!y_{n_2} \}$. In contrast to time-series classification models \citep{eddy95,kim06,FKL,quattoni10}, this allows SRNs to be used on objects from unknown classes as well. For instance, the SRN may be
trained on the signatures of a collection of people, and like any classification model, it can then be used within-domain to verify new signatures of the same people.  However, the SRN can also be used out-of-domain to verify the signatures from people that
were not present in the training set.  The SRN only needs one genuine,
verified signature to compute the similarity to a new, unknown
signature (one-shot learning).  The underlying assumption is that the inter-person variation of the
signatures is modeled well by the SRN because it was trained on signatures from many other people.

%% file: experiments.tex

We performed experiments with SRNs on three different datasets in three different learning settings: (1) within-domain similarity prediction, (2) out-of-domain similarity prediction, and (3) one-shot learning. Before presenting the setup and results of our experiments, we first introduce the three datasets below.

\subsection{Datasets}
\label{data_sets}
We performed experiments on three different datasets.

The \emph{Arabic Spoken Digit dataset} \citep{hammami09} comprises $8,800$ utterances of digits produced by $88$ different speakers. Each speaker uttered each digit ten times. The data is represented as a time series of 13-dimensional MFCCs that were sampled at $11,025$Hz and 16 bits using a Hamming window. We use two different versions of the spoken digit dataset: (1) a \emph{digit} version in which the uttered digit is the class label and (2) a \emph{voice} version in which the speaker of a digit is the class label. 

The \emph{MCYT signature dataset} \citep{ortega03} contains online signature data collected from 100 subjects. For each subject, the data comprises 25 authentic signatures and 25 skilled forgeries. The signatures are represented as time series of five features: the $x$-coordinate, $y$-coordinate, pressure, azimuth, and elevation of the pen. We consider two different versions of the dataset, namely, a version \emph{without forged} data and a version \emph{with forged} data.

The \emph{American sign language dataset} \citep{aran06} contains eight manual signs that represent different words and eleven non-manual signs such as head or shoulder motions. The data thus comprises nineteen classes. Each sign was produced five times by eight different subjects, leading to a total of $760$ samples. The time series are represented using a hand-crafted feature representation that contains a total of $77$ hand motion, hand position, hand shape, and head motion features \citep{aran06}.

\begin{table}[tb]
\setlength{\tabcolsep}{4pt}
\caption{Characteristics of the five datasets considered in our experiments: dimensionality of features, number of classes, number of samples, and the minimum, mean, and maximum length of the time series.}
\vskip 0.1in
\begin{center}
\scriptsize
\begin{tabular}{l|c|c|c|cccc}
            &      & & & \multicolumn{3}{c}{\textbf{Time series length}} \\
\textbf{Dataset}  & \textbf{Dimens.} & \textbf{Classes} & \textbf{Samples} & \textbf{Min.} & \textbf{Mean} & \textbf{Max.}\\ 
\hline
\rowcolor{gray} Arabic (digit) & $13\!\times\!2$  & 10 & 8800 & 3 & 39 & 92  \\
Arabic (voice)  & $13\!\times\!2$ & 88 & 8800 & 3 & 39 & 92  \\
\rowcolor{gray} MCYT (without forgery) & $5\!\times\!3$ & 100 & 2500 & 34 & 349 & 1161 \\
MCYT (with forgery) & $5\!\times\!3$ & 100 & 5000 & 34 & 438 & 2687 \\
\rowcolor{gray} Sign            & $77\!\times\!2$ & 19 & 40 & 760 & 112 & 198  \\
\hline
\end{tabular}
\end{center}
\label{table:data_set}
\vskip -0.1in
\end{table}
Following common practice in time series analysis, we preprocessed all three datasets by applying a sliding window (with stride $1$) to the time series, concatenating the features in the frames under the window into a single frame. 
This enriches the feature representation, making it easier for the models to capture feature gradients. For the Arabic, MCYT, and Sign datasets, we used a window size of 2, 3, and 2, respectively.    
In Table \ref{table:data_set}, the main characteristics of all five datasets are summarized.

\subsection{Experimental setup}
\label{experiment_setup}
In our experiments, the model parameters of the {\SRN}s were initialized by sampling them from a uniform distribution within an interval $[-0.1, 0.1]$. Training of the model is performed using a RMSprop \citep{tieleman12} stochastic gradient descent procedure using mini-batches of 50 pairs of time series. To prevent the gradients from exploding, we clip all gradients \citep{bengio13} to lie in the interval $[-5, 5]$. We decay the learning rate during training by multiplying it by $0.4$ every time the AUC on the validation set stops increasing.
We applied dropout on the hidden-unit activations of our model: the dropout rate was tuned to maximize the AUC on a small held-out validation set. Code reproducing the results of our experiments is available on \url{http://www.anonymized.com}.

In all experiments except for those on the MCYT (with forgery) dataset, we defined the sets of similar and dissimilar time series as suggested in Section~\ref{models_section}, that is, we define similar time series to be those with the same class label and dissimilar time series to be those with different class labels: $\mathcal{S}\!=\!\{(n_1, n_2)\!: y_{n_1}\!=\!y_{n_2} \}$ and $\mathcal{D}\!=\!\{(n_1, n_2)\!: y_{n_1}\!\neq\!y_{n_2} \}$. Herein, $y_n$ represents the class label of the time series as described in section~\ref{data_sets}. On the MCYT (with forgery) dataset, we define the positive pairs in the same way but we define the set of negative pairs $\mathcal{D}$ slightly differently: the negative pairs are pairs of a genuine signature and a forged version of the same signature. These negative pairs are more difficult to distinguish, as a result of which training on them will likely lead to better models.

We compare the performance of our SRNs with that of three variants of our model, and with three baseline models. The three variants of our model we consider are: (1) a feedforward variant of SRN-A, called SN-A, that removes all recurrent connections from the model, \emph{i.e.}, in which $\mathbf{A}\!=\!\mathbf{0}$ but which still averages the hidden representation over time; (2) a feedforward variant of SRN-L, called SN-L, that removes all recurrent connections from the model and uses the hidden representation of the last time step; and (3) a naive logistic model that removes all hidden units from the model and that predicts similarities by averaging all features over time and computing a weighted sum of the element-wise product of the resulting feature representations. These three variants of SRNs allow us to investigate the effect of the recurrent connections and non-linearities on the prediction performance of our models.

The three time-series similarity models we use as baseline models are: (1) dynamic time warping \citep{vintsyuk68}; (2) Fisher kernels~\citep{FisherKernel}; and (3) Fisher vectors~\citep{perronnin2010}. Details of these three baseline models are given below.

\begin{figure*}[t]
\vskip 0.2in
\begin{center}
\centerline{\includegraphics[width=\linewidth]{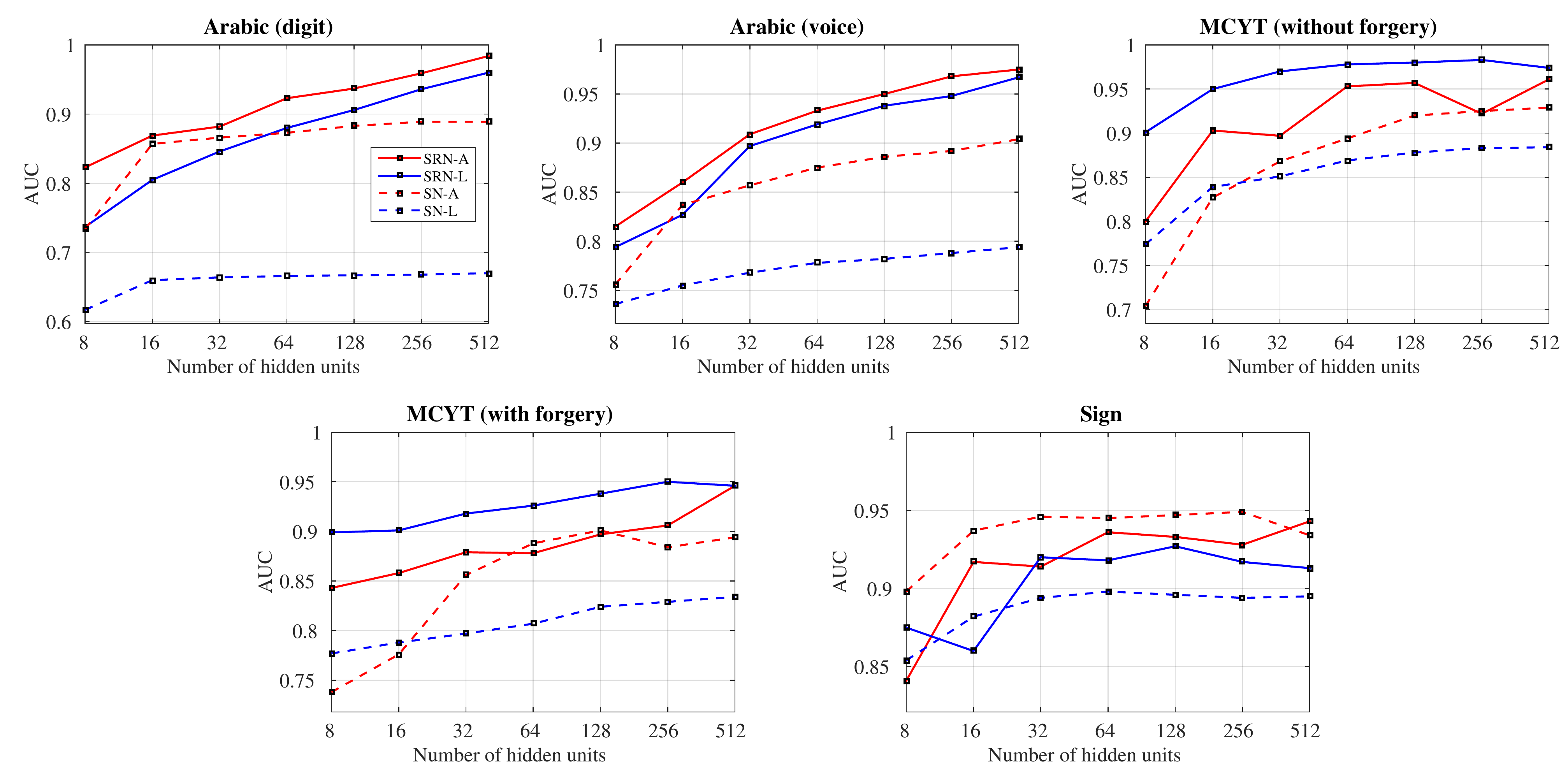}}
\caption{Area under the receiving-operator curve curve (AUC) of our two variants of Siamese Recurrent Networks (SRN-A and SRN-L) on five datasets as a function of the number of hidden units (higher is better). For reference, the performance of SRNs without recurrent connections (SNs) is also shown. All results were obtained by averaging over five repetitions. The standard deviation of the results is typically smaller than $0.01$.}
\label{fig:fixed_class_1}
\end{center}
\vskip -0.2in
\end{figure*}

\textbf{Dynamic time warping} (DTW; \citet{vintsyuk68}) measures time series similarities by aligning both time series and summing the pairwise distances between all corresponding frames, minimized over the set of all possible alignments between the two time series. An alignment is a set of (potentially many-to-many) correspondences between frames, with the restriction that correspondences cannot be crossing each other in time. DTW similarities can be computed efficiently using a dynamic-programming algorithm. Despite its simplicity, DTW has been quite successful, in particular, on problems in which the time series are already relatively well aligned and the time series show some clear salient features in time. We leave comparisons with approaches that combine dynamic time warping and metric learning \citep{Garreau14metric} to future work.

\textbf{Fisher kernels} measure the similarity between two time series by the inner product of the log-likelihood gradients that are induced by the time series with respect to the parameters of a generative model \citep{FisherKernel}. Our generative model of choice for time series is the hidden Markov model (HMM). Mathematically, we denote the gradient of the
log-likelihood $L(X^{(n)})$ of a time series $X^{(n)}$ with respect to the parameters of the HMM as $\mathbf{g}_n = \left[
\forall \theta \in \Theta: \frac{\partial L(X^{(n)})}{\partial \theta}\right]$. We define the Fisher kernel similarity $\kappa$ between two time series as an inner product between their corresponding gradients:
\begin{align}
    \kappa\left(X^{(i)}, X^{(j)}\right) = \mathbf{g}_i ^ \top \mathbf{U}^{-1} \mathbf{g}_j.
 \label{fisher_kernel}
\end{align}
Herein, the matrix $\mathbf{U}$ is the Fisher information metric, which is replaced with identity matrix $\mathbf{I}$ in our experiments. The number of hidden states of our HMMs is tuned by maximizing the AUC on a small, held-out validation set. 

\textbf{Fisher vectors} compute the same gradients $\mathbf{g}_n$ as
before, but instead of computing their inner products, we concatenate
the gradients $\mathbf{g}_i$ and $\mathbf{g}_j$ to obtain a feature representation of the time-series pair $\left(X^{(i)}, X^{(j)}\right)$. Such Fisher vector representation are commonly used in computer vision \cite{perronnin2010}. Because the concatenated Fisher vectors cannot directly measure time-series similarity, we perform 1-nearest classification on the collection of similar and dissimilar pairs to predict whether a pair of time series is similar. (In other words, the time series similarity is the negative Euclidean distance between the example and its nearest pair of similar time series in the concatenated Fisher vector space.)

\subsection{Results}
\label{results}
Below, we separately present the results for the three learning settings we considered: (1) within-domain similarity prediction, (2) out-of-domain similarity prediction, and (3) one-shot learning. We also present t-SNE visualizations of the learned time-series representations.

\subsubsection{Within-Domain Similarity Prediction}
We first evaluate the within-domain similarity prediction performance of the SRN: we randomly split the time series into a training and a test set, and we measure the ability of the models to accurately predict whether pairs of time series in the test set are similar or not in terms of the area under the receiving-operator curve (AUC). We opt for the AUC as a performance measure because it naturally deals with the potential imbalance in the sizes of $\mathcal{S}$ and $\mathcal{D}$. We refer to this experiment as \emph{within-domain} because all classes in the test data were also observed during training.  

Figure~\ref{fig:fixed_class_1} presents the within-domain similarity prediction performance of SRNs as a function of the number of hidden units in the model on five different datasets. We present results for both the variant that averages all hidden-unit activations over time (SRN-A) and the variant that uses only the hidden unit activations at the last timestep (SRN-L). The reported results were averaged over five repetitions, randomly initializing the parameter of the models in each repetition.  The figure also reports the performance of models without recurrent connections, called a Siamese network (SN, where SN-A is a Siamese network with averaged hidden activations and SN-L is a network that uses the last time step activations). From the results presented in Figure~\ref{fig:fixed_class_1}, we make three main observations. 

\begin{figure}[t]
\vskip 0.2in
\begin{center}
\centerline{\includegraphics[width=1.0\columnwidth]{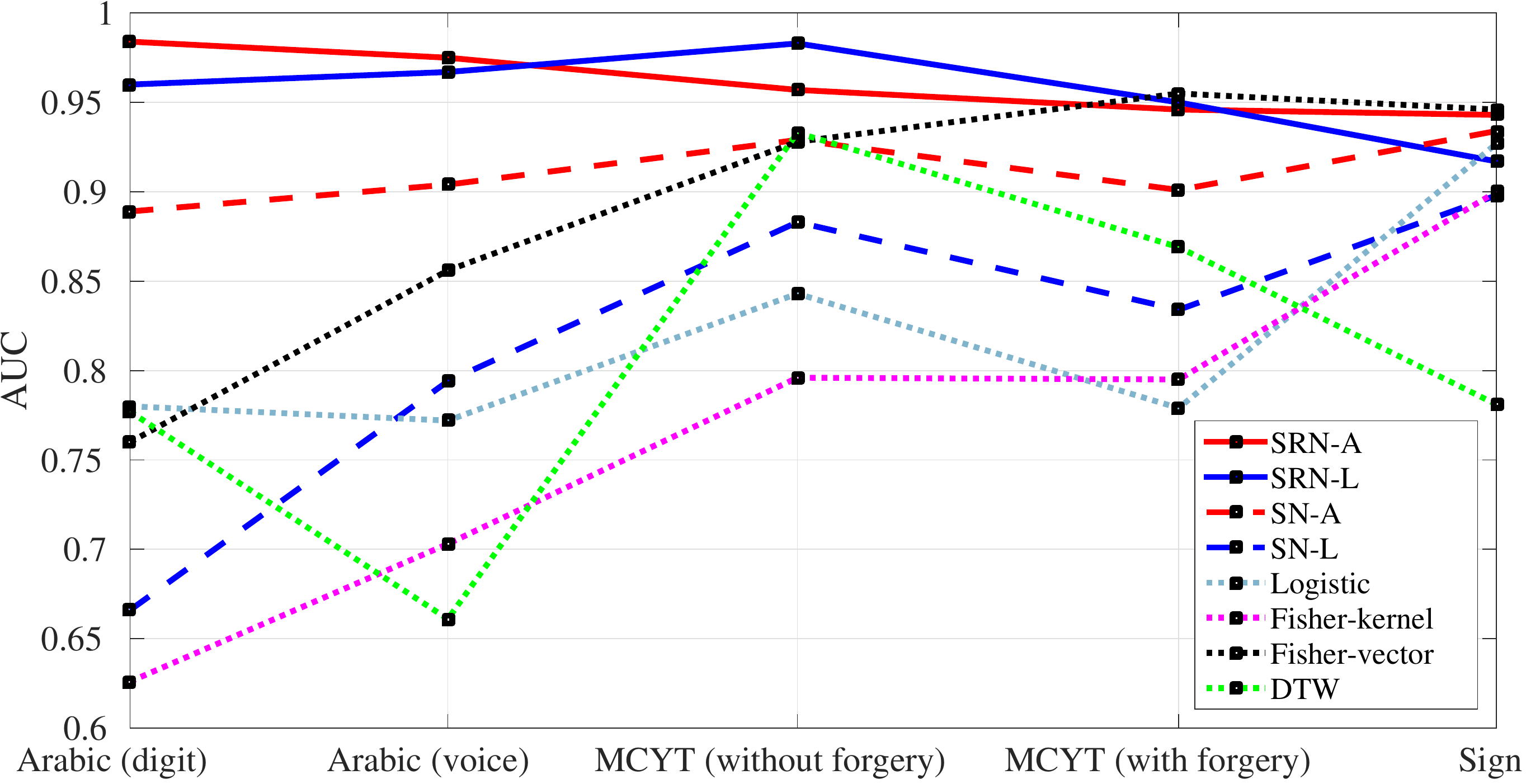}}
\caption{Area under the receiving-operator curve curve (AUC) on five different datasets using eight different time-series similarity learning models in a within-domain similarity prediction setting (higher is better). See text for details.}
\label{fig:model_comparison}
\end{center}
\vskip -0.2in
\end{figure} 

First, the results show that the performance of SRNs tends to increase with the number of hidden units, in particular, on challenging datasets such as the Arabic speech datasets. This shows that SRNs effectively use the additional capacity that is provided by additional hidden units to learn more informative features for the time-similarity measurements. In our experiments, we did not observe much overfitting, although overfitting is likely to occur when the number of hidden units is increased much further.

Second, we observe that there is no clear winner between averaging hidden unit activations over time (SRN-A) and using the activations at the last timestep (SRN-L). This suggests that the recurrent networks in the SRN-L models are at least partly successful in remembering relevant features over time.

Third, we observe that the recurrent connections in the SRN models are, indeed, helpful: the SRN models outperform their counterparts without recurrent connections (SNs) in nearly all experiments\footnote{It should be noted that because we preprocess the time-series data by windowing features, the SN is actually a convolutional network that is very similar to the time-delay neural networks of \citet{bromley93}.}. This result underlines the hypothesis that recurrent connections can preserve features relevant for time-series similarity computations over time. Somewhat surprisingly, the performance of the SN-L models is not as bad as one may expect. It should be noted that the windowing of features makes the feature representation of the last timestep richer, which is sufficient to obtain acceptable performances on some of the datasets.

\begin{table*}[tb]
\caption{Area under the receiving-operator curve curve (AUC) of eight time-series similarity models on five datasets in an out-of-domain similarity prediction setting (higher is better). The standard deviation of the five repetitions we performed is typically smaller than $0.01$. The best performance per dataset is boldfaced. See text for details.}
\vskip 0.1in
\setlength{\tabcolsep}{4pt}
\begin{center}
\renewcommand\arraystretch{1.2}
\begin{tabular}{lcc|cccccccc}
\hline
\multirow{2}{*}{\textbf{Dataset}} & \multirow{2}{*}{\textbf{\parbox{1.2cm}{\centering Training\\classes}}} &
\multirow{2}{*}{\textbf{\parbox{1.2cm}{\centering Test\\classes}}} & \multicolumn{8}{c}{\textbf{Model}}
\\
 & & &  \textbf{SRN-A} & \textbf{SRN-L} & \textbf{SN-A} & \textbf{SN-L} & \textbf{Logist.} & \textbf{DTW} & \textbf{Fisher K.} & \textbf{Fisher V.}\\ 
\hline
\rowcolor{gray}Arabic (digit) & 1-7 & 8-10 &0.681  & 0.714 & \textbf{0.768} & 0.539& 0.761& 0.725 &0.600 &0.561    \\
Arabic (voice)  & 1-60 & 61-88 & \textbf{0.849} &0.788 & 0.802 &0.684 &0.730 &0.640 & 0.698 & 0.630\\
\rowcolor{gray} MCYT (without forgery) & 1-70 & 71-100& 0.914 & 0.920& 0.816&0.760 & 0.824& \textbf{0.952}& 0.752&0.844 \\
MCYT (with forgery)  & 1-70 & 71-100 & 0.888 &0.876 & 0.828 &0.668& 0.782&\textbf{0.894} &0.805 &0.813 \\
\rowcolor{gray} Sign  & 1-14 & 15-19 & \textbf{0.862} &0.670 & 0.748 & 0.565 & 0.836 & 0.729& 0.770&0.566 \\
\hline
\end{tabular}
\end{center}
\label{table:verification}
\end{table*}

\textbf{Comparison with baseline models.} Next, we compare the performance between of SRNs with the naive logistic model and three other baseline time-series similarity learning models: (1) dynamic time warping, (2) Fisher kernels, and (3) Fisher vectors (see section ~\ref{experiment_setup} for details). We used the same experimental setup as in the previous experiment, but we tuned the main hyperparameters of the models (the number of hidden units in SRNs and SNs; the number of HMM hidden states for Fisher kernels and Fisher vectors) on a small held-out validation set. Figure~\ref{fig:model_comparison} presents the results of these experiments.

The results of these experiments show that, indeed, the SRN can be a very competitive time-series similarity model, even when trained on relatively small datasets. In particular, SRNs substantially outperform the baselines models on the Arabic (digit), Arabic (voice), and MCYT (without forgery) datasets. On most datasets, the Fisher vectors are the best baseline model (they perform substantially better than standard Fisher kernels), which is line with results in the literature \citep{perronnin2010}. The naive logistic model performs substantially worse than the SRN models, which suggests that hidden units are essential in solving difficult similarity assessment problems.

Dynamic time warping (DTW) performs reasonably well on relatively simple datasets such as the Sign dataset, but its performance deteriorates on more challenging datasets in which the similarity labels are not aligned with the main sources of variation in the data, such as the Arabic (voice) dataset: the main sources of variation in this dataset are likely due to the differences in the digits being uttered, whereas the similarity labels we are interested in concern the speaker of the digit and not the digit itself. DTW (as well as Fisher vectors and kernels) cannot exploit this information, which explains its inferior performance on the Arabic (voice) dataset.

\subsubsection{Out-of-Domain Similarity Prediction}
\label{out-of-domain}
In the next set of experiments, we measure the performance of SRNs on \emph{out-of-domain} similarity prediction: we use the same experimental setup as before, however, we split the training and test data in such a way that the set of class labels appearing in the training set and the set of class labels appearing in the test set are disjoint. This is a more challenging learning setting, as it relies on the time-series similarity model exploiting structure that is shared between classes in order to produce good results. We obtain the test data by selecting 3 out of 10 classes on the Arabic (digit) dataset, 28 out of 88 classes on the Arabic (voice) dataset, 30 out of 100 classes on the MCYT datasets, and 5 out of 19 classes on the Sign dataset. As before, we measure the performance of our models in terms of AUC, and we tune the hyperparameters of the models on a validation set. The results of these experiments are presented in Table~\ref{table:verification}.

From the results presented in the table, we observe that the strong performance of SRNs on difficult datasets such as the Arabic (voice) datasets generalizes to the out-of-domain prediction setting. This suggests that, indeed, the SRN models are able to learn some structure in the data that is shared between classes. On the (much smaller) MCYT datasets, however, dynamic time warping outperforms SRNs. Most likely, this result is caused by the SRNs (which have high capacity) overfitting on the classes that are observed during training.
     
\subsubsection{One-Shot Learning}
To further explore the potential of SRNs in out-of-domain settings, we
performed an experiment in which we measured the performance of SRNs in
one-shot learning. We adopt the same dataset splits as
in~\ref{out-of-domain} to obtain train and test data. On the training
data, we train the SRNs to learn a similarity measure for time series. This similarity measure is used
to train and evaluate a nearest-neighbor classifier on the test set. We use only a single time series per class from the test set
to train the nearest-neighbor classifier, and use the remaining time series in the test set for evaluation. We measure the classification accuracy using leave-one-per-class-out validation.

\begin{table}[!tb]
\caption{Classification accuracy of one-shot learning models of an 1-nearest neighbor classifier using three different similarity measures on four different datasets (higher is better). The best performance per dataset is boldfaced. See text for details.}
\vskip 0.2in
\begin{center}
\begin{tabular}{l|cccc}
\textbf{Dataset} & \textbf{SRN-A} & \textbf{SRN-L} & \textbf{DTW}\\ 
\hline
\rowcolor{gray} Arabic (digit) & 0.618 &0.613 &\textbf{0.801} \\
Arabic (voice)  & \textbf{0.273} & 0.228 &0.151 \\
\rowcolor{gray} MCYT (without forgery) & 0.418 & 0.548& \textbf{0.913} \\
Sign  & \textbf{0.599} & 0.381&0.531 \\
\hline
\end{tabular}
\end{center}
\label{table:one-shot}
\vskip -0.2in
\end{table}

The results are presented in Table~\ref{table:one-shot}. For datasets that have clear salient features, like the MCYT, and to a lesser degree the Sign dataset, DTW performs well. For more complex data, the SRN performs well provided that sufficient training data is available. For the Arabic (digit) dataset, the seven classes used in training are insufficient for the SRN, and the SRN overfits on those classes. On the Arabic (voice) dataset 60 classes are available, which allows the SRN to fully exploit its potential.

\subsubsection{Visualizing the Representations}
\begin{figure*}[!t]
\begin{flushleft}
$\begin{array}{ccc}
  \includegraphics[width=0.30\linewidth]{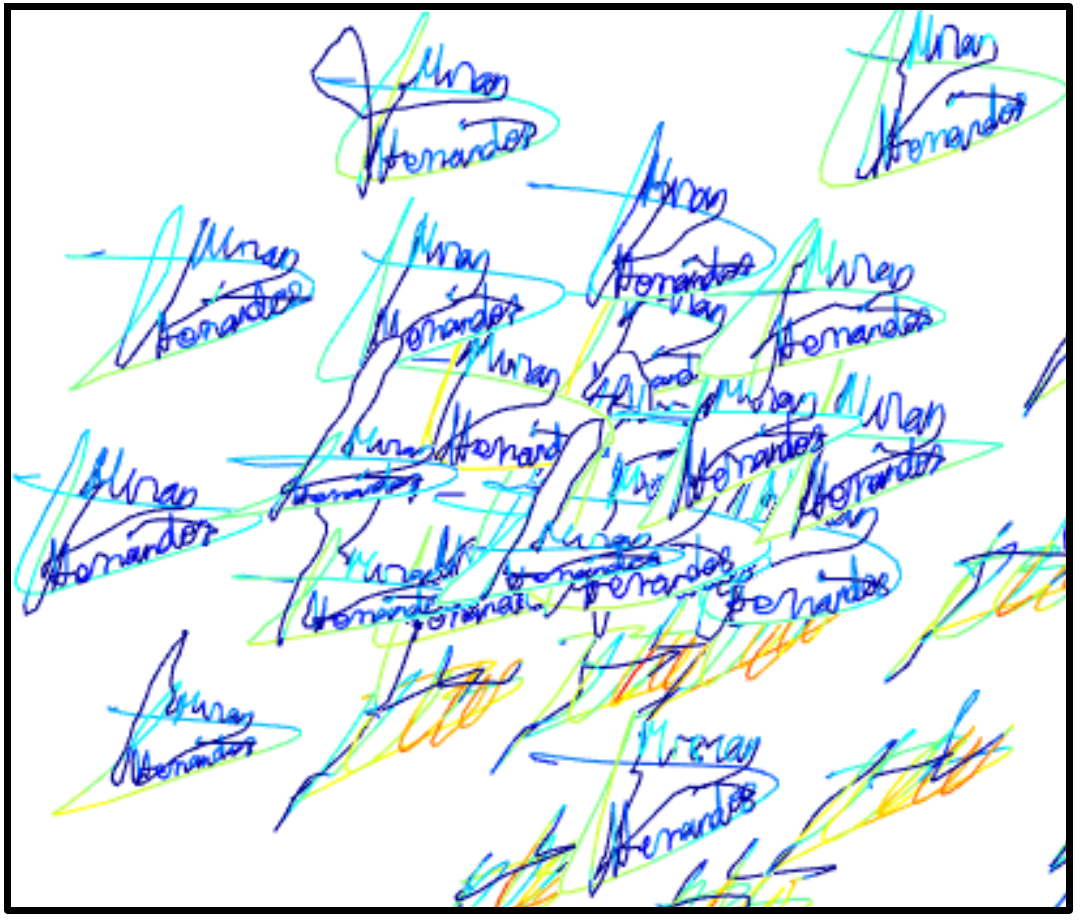} &
\includegraphics[width=0.33\linewidth]{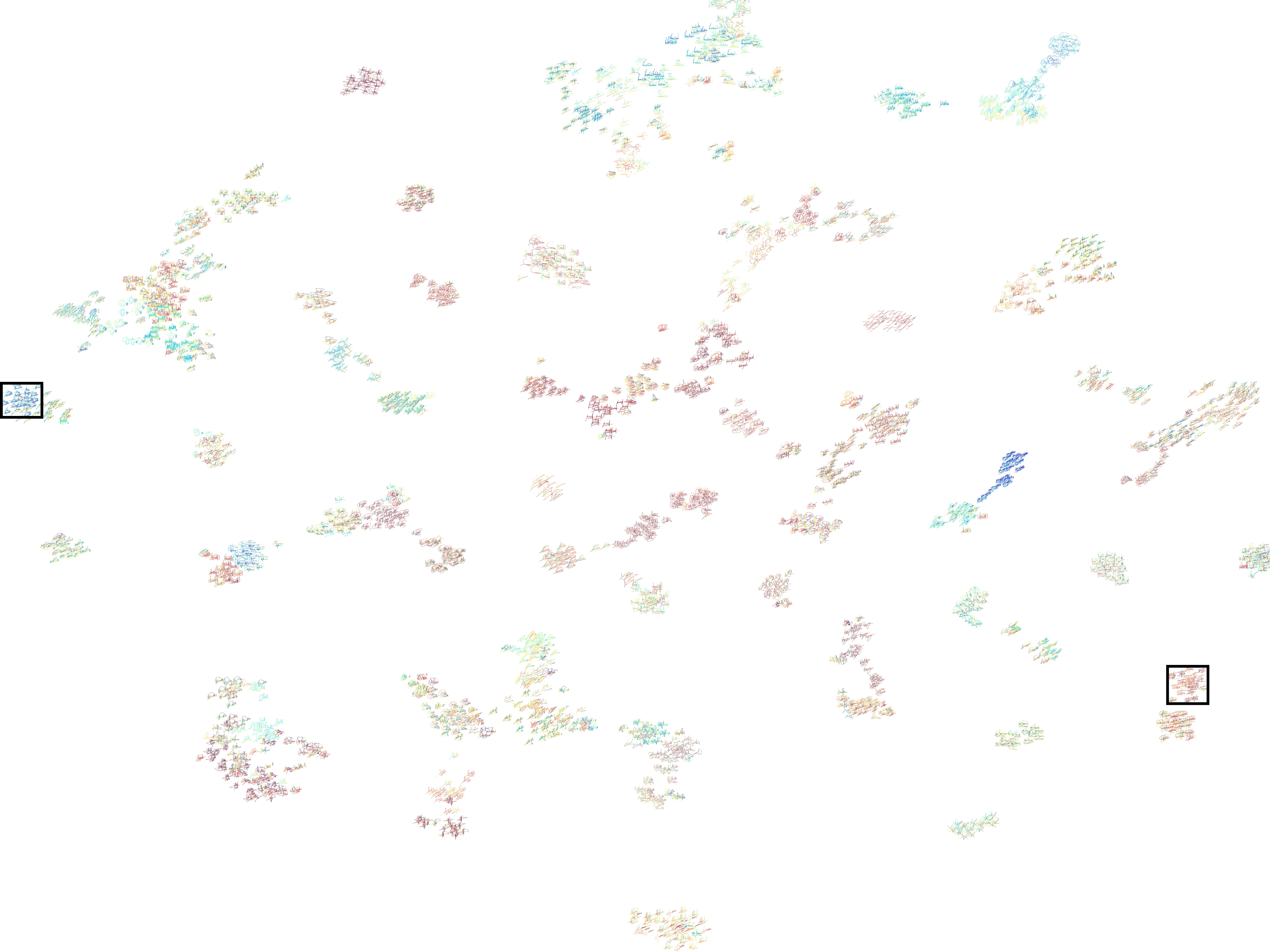} &  
   \includegraphics[width=0.33\linewidth]{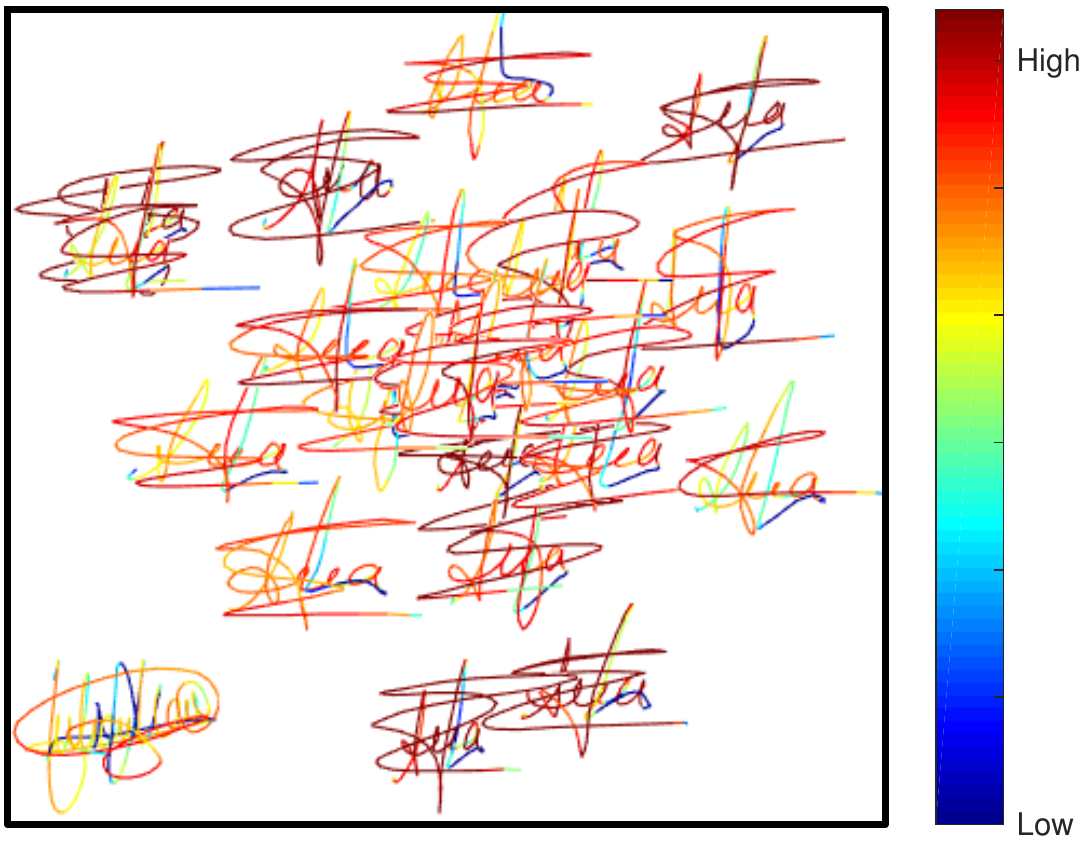} 
\end{array}$
\end{flushleft}
   \caption{t-SNE map of the $2,500$ signatures in the MCYT test set (100 subjects) data based on embeddings computed by an SRN-L. The signatures were drawn by integrating the pen movements over time, and colors indicate the pen pressure (red indicates high pressure and blue indicates low pressure). A full-resolution version of this map is presented in the supplemental material.}
\label{fig:t-SNE_MCYT}
\end{figure*}

The one-shot learning experiment presented above exploits an interesting property of the SRN model, namely, that it learns a single embedding for a time series. An advantage of this is that the resulting time-series embeddings can be used in a wide variety of other learning algorithms that operate on vectorial data, such as alternative classification techniques, clustering models, \emph{etc.} To obtain more insights into what the SRN models have learned, we apply t-SNE \citep{vandermaaten08} on embeddings obtained by a SRN-L on the MCYT (without forgery) test set.  Figure~\ref{fig:t-SNE_MCYT} shows a map of the 2500 signatures in the test set; the signatures were drawn by integrating the pen movements over time. The color indicates the pen pressure. We refer the reader to the supplemental material for a full-resolution version of this map. The t-SNE visualization shows that, indeed, the SRN-L is capable of grouping similar signatures together very well. 

\begin{figure}[ht]
\begin{center}
\footnotesize
$\begin{array}{c}
  \includegraphics[width=0.61\linewidth]{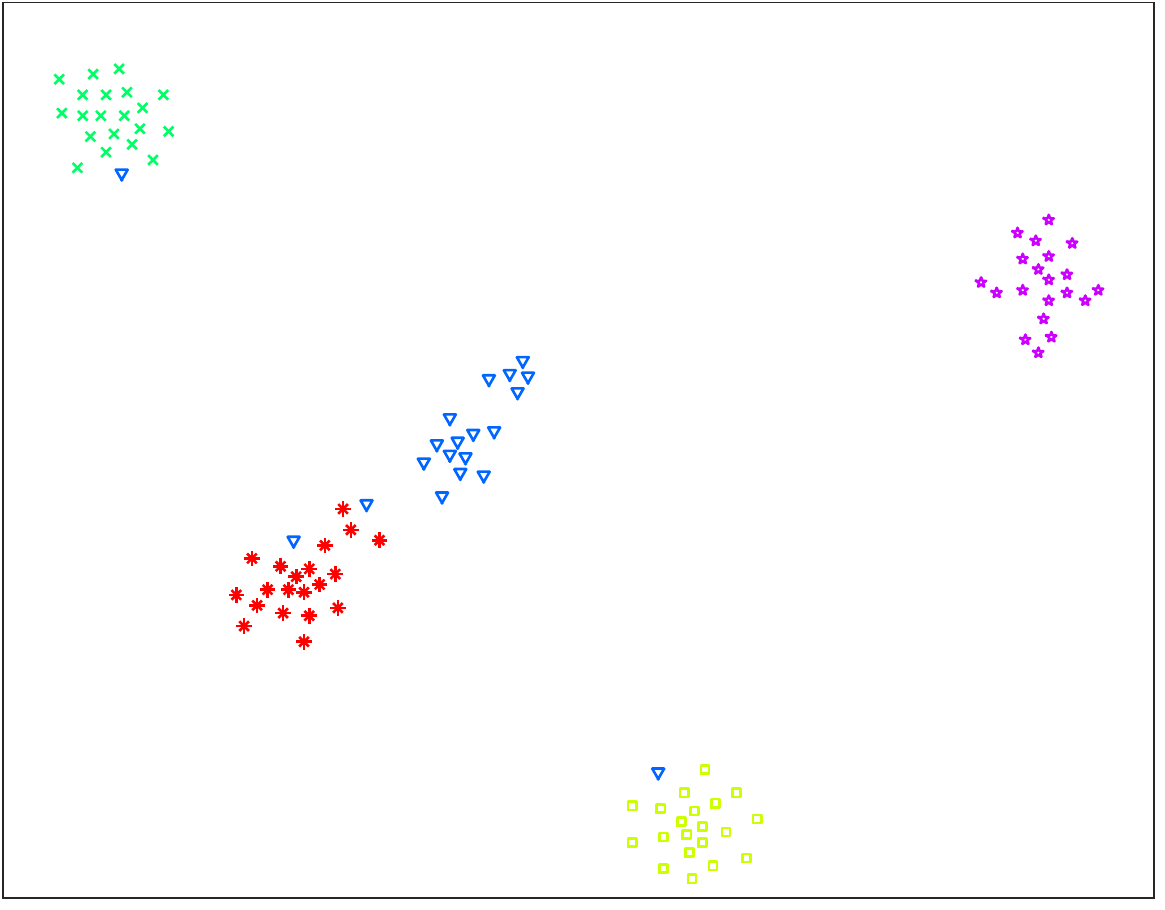} \\ \text{(a) SRN-L.} \\ \\
  \includegraphics[width=0.61\linewidth]{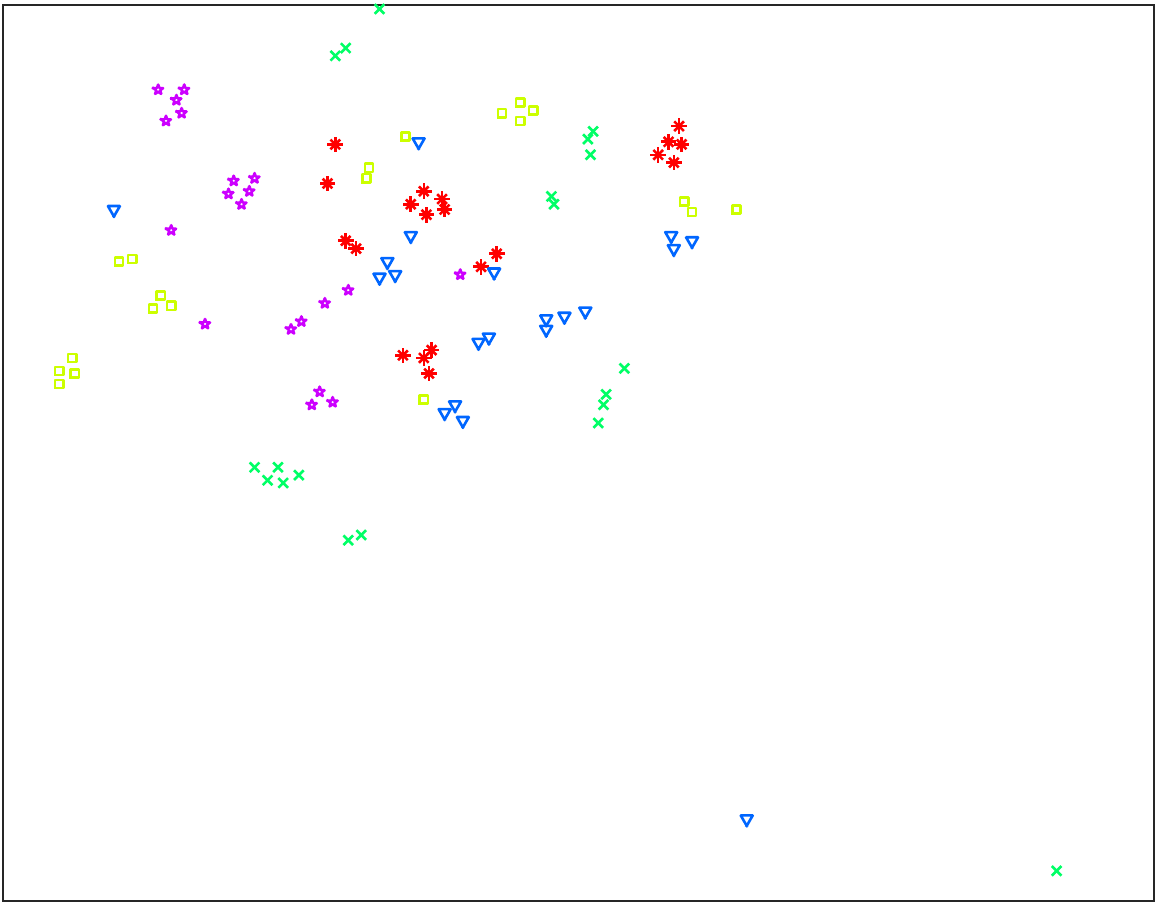} \\ \text{(b) DTW.} 
\end{array}$
\end{center}
   \caption{t-SNE maps of the Arabic (voice) test data from five randomly selected classes, constructed based on (a) siamese recurrent network (last timestep) embeddings of the time series and (b) pairwise similarities computed using dynamic time warping.}
\label{fig:t-SNE_arabic_voice_mixed}
\end{figure}

In Figure~\ref{fig:t-SNE_arabic_voice_mixed}, we show a t-SNE map of the Arabic (voice) test set constructed on SRN-L embeddings. For comparison, we also show a t-SNE map of the same data, based on pairwise distances computed with DTW. The two maps clearly show the potential advantage of SRN: it has used the supervised similarity information to group all the utterances corresponding to a single class together, something that DTW is unable to do due to its unsupervised nature.

%% file: conclusions.tex
We have investigated models for learning similarities between time series based on supervised information. Our study shows that a combination of ideas from metric learning and deep time-series models has the potential to improve the performance of models for time-series classification, retrieval, and visualization. The proposed siamese recurrent networks (SRNs) are particularly effective compared to alternative techniques in settings in which the similarity function that needs to be learned is complicated, or when the number of labeled time series for some of the classes of interest is limited. When a reasonably large collection of examples of similar and dissimilar time series is available to train the models, the siamese recurrent networks can produce representations that are suitable for challenging problems such as one-shot learning or extreme classification of time series. This result is in line with earlier results for siamese convolutional networks by, for instance, \citet{kamper16}.

This study is an initial investigation into learning similarities between time series, and we foresee several directions for future work.  In particular, we intend to explore variants of our model architecture: (1) that employ a bilinear model to measure the similarity of the RNN representations; (2) that employ long-term short-term units \citep{hochreiter97} or gated recurrent units \citep{cho14} instead of the simple rectified linear units we are currently using; (3) that employ multiple layers of recurrent units; and (4) that have a tree structure or generic (planar) graph structure instead of the current sequential structure.  The latter extension would make our models applicable to problems such as molecule classification \citep{riesen08}. We also plan to explore improvements to our learning algorithm. In particular, our current implementation selects negative pairs of time series in a somewhat arbitrary manner: in all our experiments, we select negative examples uniformly at random for the set of all candidate negative pairs. We plan to investigate approaches that perform a kind of ``hard negative mining'' during learning, akin to some modern metric learning \citep{weinberger09distance} and multi-modal learning \citep{weston11} approaches. We also plan to study applications of SRNs in, for instance, learning word-discriminative acoustic features \citep{synnaeve2014phonetics}.